\documentclass[10pt,twocolumn,letterpaper]{article}

\usepackage[pagenumbers]{cvpr} 
\usepackage{multirow}
\usepackage{xcolor}
\definecolor{cvprblue}{rgb}{0.21,0.49,0.74}
\usepackage[pagebackref,breaklinks, colorlinks,allcolors=cvprblue]{hyperref}

\title{Joint Post-Training Quantization of Vision Transformers with Learned Prompt-Guided Data Generation}

\author{Shile Li\\
vivo Tech Research GmbH\\
{\tt\small shile.li@vivo.com}
\and
Markus Karmann\\
Torr Vision Group\\Universtiy of Oxford\\
{\tt\small markus.karmann@eng.ox.ac.uk}
\and   
Onay Urfalioglu \\
vivo Tech Research GmbH\\
{\tt\small onay.urfalioglu@vivo.com}
}

\begin{document}
\maketitle

\begin{abstract}
We present a framework for end-to-end joint quantization of Vision Transformers trained on ImageNet for the purpose of image classification. 
Unlike prior post-training or block-wise reconstruction methods, we jointly optimize over the entire set of all layers and inter-block dependencies without any labeled data, scaling effectively with the number of samples and completing in just one hour on a single GPU for ViT-small.
We achieve state-of-the-art W4A4 and W3A3 accuracies on ImageNet and, to the best of our knowledge, 
the first PTQ results that maintain strong accuracy on ViT, DeiT, and Swin-T models under extremely low-bit settings (W1.58A8), demonstrating the potential for efficient edge deployment.
Furthermore, we introduce a data-free calibration strategy that synthesizes diverse, label-free samples using Stable Diffusion Turbo guided by learned multi-mode prompts.  
By encouraging diversity in both the learned prompt embeddings and the generated image features, our data-free approach achieves performance on par with real-data ImageNet calibration and surpasses simple text-prompt baselines such as 
\textit{``a \texttt{<adjective>} photo of \texttt{<adjective> <cls>}''}.

\end{abstract}

\section{Introduction}
\label{sec:intro}

\begin{figure}[t]
    \centering
    \includegraphics[width=\linewidth]{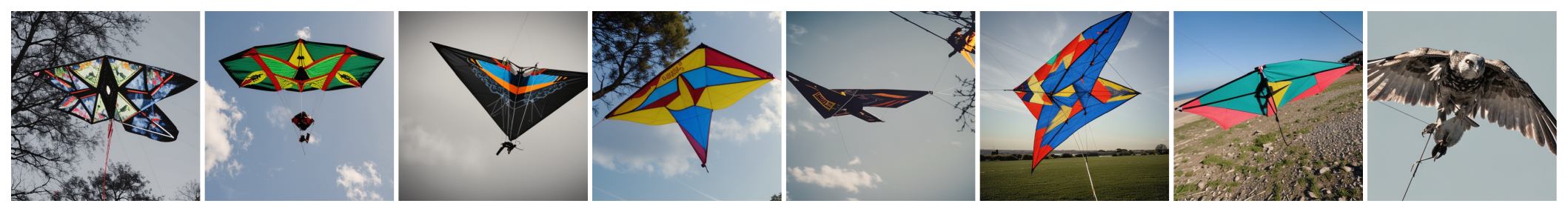}
    \vspace{-1pt}
    \includegraphics[width=\linewidth]{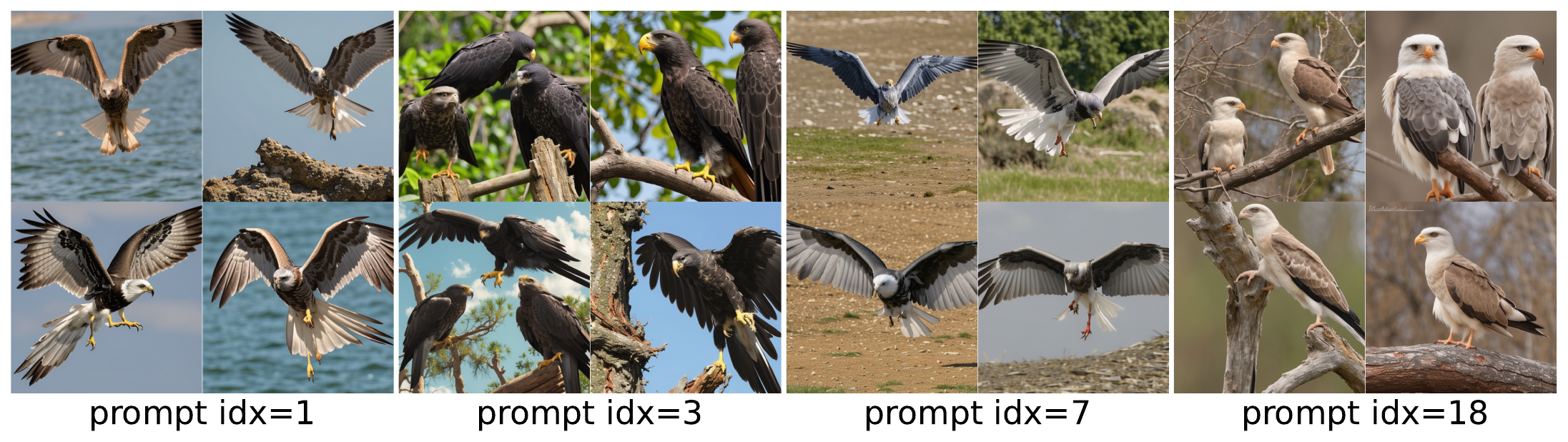}
    \includegraphics[width=\linewidth]{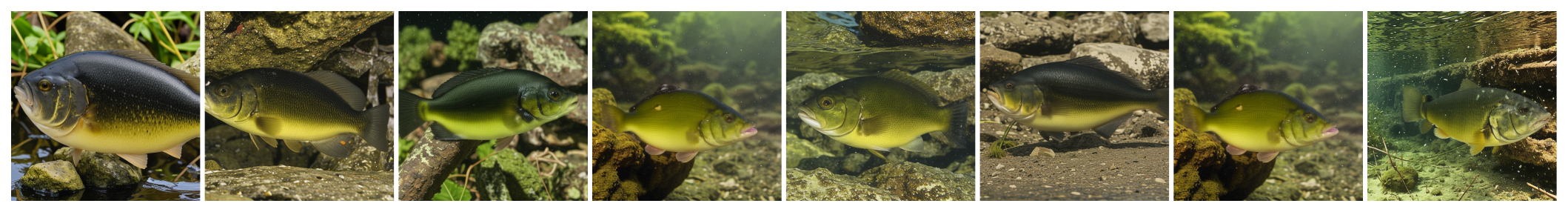}
    \vspace{5pt}
    \includegraphics[width=\linewidth]{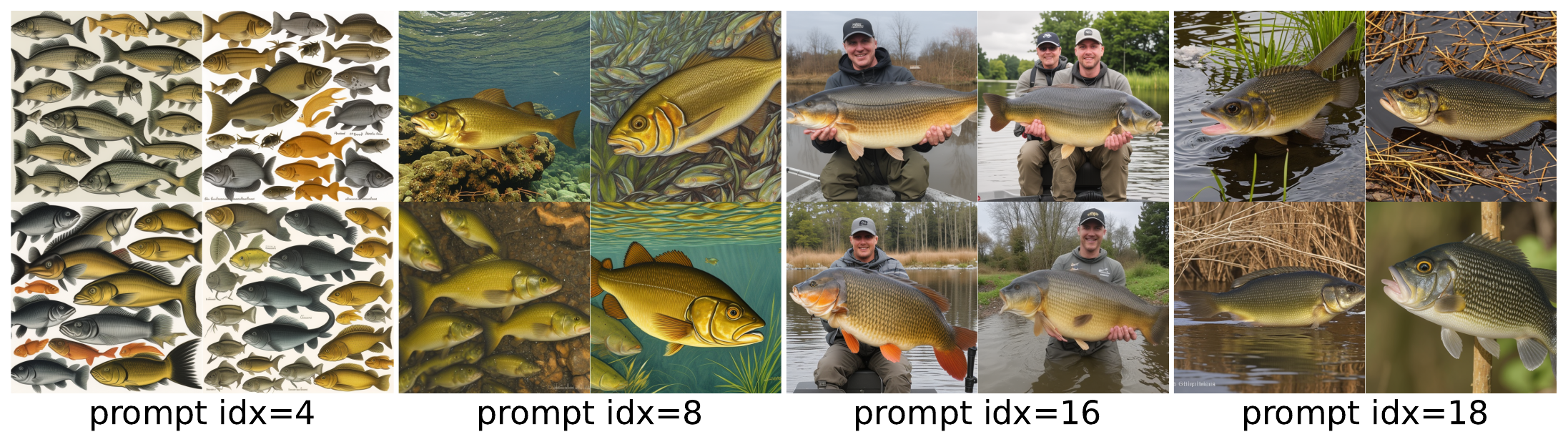}
    \includegraphics[width=\linewidth]{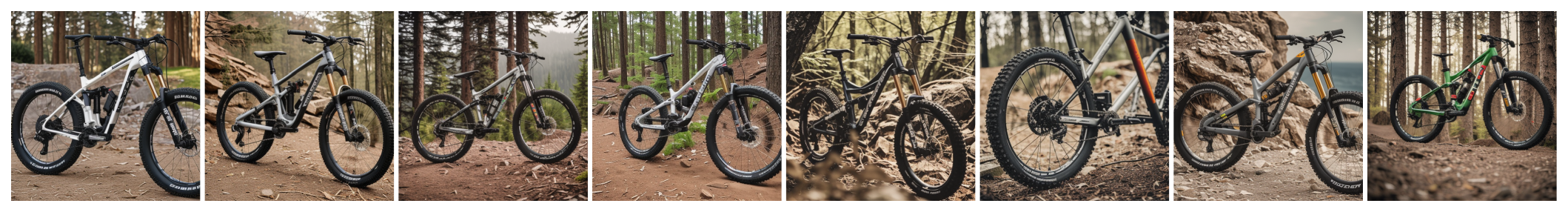}
    \vspace{5pt}
    \includegraphics[width=\linewidth]{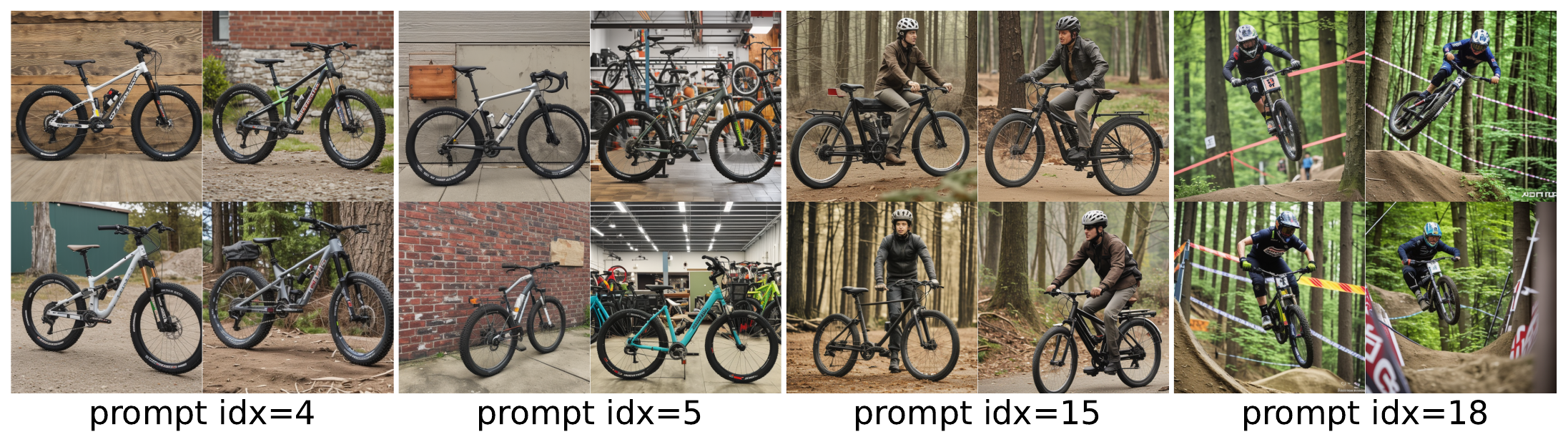}
    \caption{
    Our PTQ framework also supports a fully data-free mode using image synthesis via Stable Diffusion Turbo.  
    For each ImageNet class (\textit{kite}, \textit{tench}, and \textit{mountain bike}),  
    \textbf{top:} images from simple text prompts show limited diversity and occasional semantic errors 
    (e.g., \textit{kite} as a toy rather than a bird).  
    \textbf{bottom:} our learned multi-mode prompts (4 of 20 per class shown) generate semantically correct and diverse samples in layout, background, and style.
    These synthetic images are used for calibration in our data-free quantization pipeline.
    }
    \vspace{-10pt}

    \label{fig:qualitative_raw_vs_learned}
\end{figure}

Vision Transformers (ViTs) have achieved remarkable success across a wide range of visual recognition tasks \cite{dosovitskiy2020image}, benefiting from large-scale pretraining and strong modeling capacity. However, these models are demanding in terms of computation and memory, making them difficult to deploy on resource-limited devices or in real-time applications \cite{chen2021crossvit}.
Model quantization provides an attractive solution by reducing the precision of weights and activations, significantly decreasing model size and inference cost \cite{nagel2020up, wu2024adalog, li2023repq, lin2021fq}. Yet, conventional quantization-aware training (QAT) requires labeled data and long fine-tuning cycles, which are both time- and energy-intensive \cite{krishnamoorthi2018quantizing, li2022qvit}.
As a result, post-training quantization (PTQ), which avoids full retraining has gained attention as a more practical alternative \cite{li2023repq, wu2024adalog, wu2025aphq}. However, applying PTQ to ViTs remains challenging due to their inter-layer and inter-block dependencies, sensitivity to non-uniform channelwise-activation scaling, and complex attention mechanisms that make independent layer calibration suboptimal \cite{li2021brecq, ding2023cbq}.

Most PTQ methods developed for convolutional networks perform layer-wise calibration or block-wise reconstruction to minimize output mismatch \cite{li2021brecq, liu2023pd}. When extended to transformers, such strategies often break down because attention layers introduce strong interdependencies across blocks, making isolated reconstruction insufficient.
Recent approaches such as SmoothQuant, RepQ-ViT \cite{xiao2023smoothquant,  shao2023omniquant, li2023repq}, and other block reconstruction frameworks attempt to mitigate this via channel rescaling or sequential reconstruction. However, these methods still ignore global correlations and can become trapped in overly rigid local reconstruction objectives, preventing effective cross-block adaptation.
Although many of these methods claim efficiency by requiring only 1k calibration samples, their performance saturates quickly and fails to improve with larger calibration sets, indicating a limitation of block-wise design.
Moreover, to the best of our knowledge, no prior PTQ method achieves extreme low-bit quantization (i.e., ternary weights) on Vision Transformers. Some QAT-based works can reach such precision but at the cost of full or longer retraining with labeled data \cite{li2022qvit}.
To address these limitations, we propose an end-to-end PTQ optimization framework for the ImageNet classfication task, that jointly optimizes all transformer blocks without using labels and completes in only 1-2.5 hours on a single GPU.

Unlike previous block-wise reconstruction methods, our approach models inter-block dependencies and learns layer-wise rescaling, clipping, and channel-wise quantization parameters under a unified objective.
By formulating the PTQ optimization globally rather than sequentially, our method leverages the representational redundancy between blocks, improving both optimization stability and low-bit accuracy.
The optimization is computationally efficient and achieves state-of-the-art results at W4A4 precision, as well as robust W1.58A8 quantization.
Furthermore, we show that the framework benefits from larger calibration sets when available, but remains effective even with minimal or entirely synthetic data.
This enables high-accuracy quantization of ViTs under both data-scarce and data-free scenarios.

To completely remove dependence on real calibration data, we further propose a data-free calibration strategy built upon Stable Diffusion Turbo \cite{sauer2024adversarial}.
A straightforward approach would be to use hand-crafted text prompts such as “a \texttt{<adjective>} photo of \texttt{<class>}”, where the adjective can vary across different styles (e.g., “a close-up photo”, “a detailed photo”) \cite{li2024genq, park2025enhancing}. However, this baseline prompt-based method often produces highly similar images with limited diversity, which restricts its effectiveness for quantization. Moreover, text ambiguity in certain class names (e.g., “crane”: bird or construction machine, “kite”: bird or toy, “drake”: artist or duck) further reduces the consistency and representativeness of the generated samples.
In this work, instead of relying on manually designed prompts, we learn class-specific prompts automatically, not just one per class, but multiple diverse prompt embeddings to improve coverage of the visual space.
Training is guided by the classification signal from a pretrained full-precision ViT, without using any real images. To encourage diversity, we introduce complementary objectives in the text embedding space, the generated image space, and the ViT feature space, ensuring that the learned prompts capture distinct semantics and spatial layouts.
Specifically, we learn the token embeddings of multiple prompts for each class by optimizing them under this classification guided objective. These multiple modes drive the diffusion model to synthesize diverse and semantically rich calibration images, providing variations in object appearance, layout, and illumination.
By enforcing intra-class diversity and feature-level orthogonality, the generated samples effectively approximate the activation distribution of real ImageNet data.
Empirically, this synthetic calibration performs on par with real-data calibration and outperforms prompt-based baselines that use simple templates such as \textit{“a \texttt{<template>} photo of \texttt{<cls>}”}.
This demonstrates that image samples from learned generative priors can effectively replace real data for Vision Transformer quantization.

In summary, our main contributions are as follows:

\begin{itemize}
    \item An end-to-end PTQ framework for Vision Transformers that jointly optimizes all quantization parameters across blocks and layers without any labeled data.
    \item A generative, data-free calibration strategy that leverages ViT classification signals to train multi-mode concept embeddings within Stable Diffusion Turbo, achieving calibration performance comparable to real data.
    \item Extensive experiments demonstrating state-of-the-art quantization performance on ViT, DeiT, and Swin-T models under low bit settings, each requiring approximately 1-2.5 hours on a single GPU.   
\end{itemize}

    

\section{Related Works}
\label{sec:related}

\paragraph{Post-Training Quantization (PTQ)}
aims to compress full-precision models into low-bit formats without full retraining \cite{krishnamoorthi2018quantizing, esser2019learned}. 
This paradigm has achieved remarkable success on CNNs, mainly because convolutional weights and activations often have approximately Gaussian or symmetric distributions, making them easier to quantize. 
Representative PTQ methods such as DFQ \cite{nagel2019data}, AdaRound \cite{nagel2020up}, and BRECQ \cite{li2021brecq} demonstrate that careful layer-wise calibration or reconstruction can nearly close the gap with quantization-aware training (QAT). 
However, these approaches rely on structural and statistical assumptions that do not hold for ViTs, leading to a significant accuracy drop when directly applied to ViTs.

The ViT \cite{dosovitskiy2020image} has become the de facto backbone in modern vision models.  However, its unique architectural and statistical properties make quantization more challenging than in CNNs. 
First, the activation distributions in ViTs, especially after the softmax layers, are highly non-Gaussian and are partly heavy-tailed, containing large outlier values
\cite{liu2021post, yuan2022ptq4vit}. 
Second, there is a large inter-channel magnitude variation in the activations, which breaks the uniform scale assumption that standard quantizers rely on. 
Consequently, directly applying CNN-oriented PTQ results in severe degradation, especially under low-bit settings such as W4A4 or W3A3.
To alleviate these issues, several works have explored QAT-based solutions. 
PackQ-ViT \cite{dong2023packqvit} and Q-ViT \cite{li2022qvit} introduce learnable quantization parameters during fine-tuning and achieve strong accuracy under 4-bit settings. 
However, these methods require end-to-end retraining with labeled data, which is resource-intensive and unsuitable for data-free or resource-limited scenarios.

To better fit ViT activation distributions, FQ-ViT \cite{lin2021fq} introduces logarithmic (power-of-two) quantization to allocate denser levels for near-zero-magnitude activations while maintaining larger step sizes for outliers. 
Although effective, such non-uniform quantizers often lack hardware compatibility and require custom inference kernels. 
In contrast, our method adheres to standard uniform quantizers but introduces learnable rescaling factors to fully exploit their capacity.
RepQ-ViT \cite{li2023repq} proposes scale reparameterization to better align weight and activation ranges, but its per-layer grid search is limited to a small calibration set and ignores cross-block dependencies. 
As a result, its performance drops sharply at W4A4 and below, and it further relies on a non-standard $\log\sqrt{2}$ quantizer, complicating hardware deployment.  
APHQ-ViT \cite{wu2025aphq} mitigates activation nonlinearity by replacing \texttt{GELU} with \texttt{ReLU} and learning a small MLP network to reconstruct post-ReLU activations.  
FIMA-Q \cite{wu2025fima} proposes an efficient Fisher Information Matrix approximation to guide quantization via block-wise reconstruction loss, pushing ViT PTQ toward W3A3 and W4A4 with notable improvement.  
However, the block-level independence neglects inter-block compensation, leaving further potential unexploited.  
To the best of our knowledge, no prior work has yet demonstrated successful post-training quantization at sub-2-bit (e.g., W1.58A8) precision.

\paragraph{Data-Free Quantization (DFQ)}
aims to compress a full-precision model without access to its original training data, thus enabling deployment under data-privacy or data-unavailability constraints.  
Early approaches such as Nagel et al.\,\cite{nagel2019data} perform weight equalization and bias correction to balance layer statistics and approximate model behaviour with synthetic inputs.  
Subsequent generative methods, such as Xu et al.\,\cite{xu2020generative}, introduce a generator trained to match internal activation statistics, thereby producing synthetic calibration samples more consistent with the original data distribution.  

Applying DFQ to Vision Transformers (ViTs) is substantially more difficult due to the absence of batch-normalization statistics, the presence of large self-attention modules, and highly variant activation distributions.  
Li et al.\,\cite{li2022patch,li2023psaq} address these issues by defining a patch-similarity metric to guide synthetic sample generation and improve the representativeness of pseudo data for ViTs.  
Ramachandran et al.\,\cite{ramachandran2024clamp} further enhance semantic richness through patch-level contrastive learning, yielding more informative synthetic features for calibration.  

While most existing DFQ methods rely on Gaussian noise or simple synthetic reconstructions, recent work recognizes the need for semantically meaningful and diverse calibration data.  
Li et al.\,\cite{li2024genq} propose to employ a text-to-image diffusion model to synthesize calibration images guided by class-level text prompts, followed by filtering mechanisms that align the generated distribution with real-image statistics.  
More recently, Park et al.\,\cite{park2025enhancing} introduce a mixup-class prompting strategy that fuses multiple class labels at the prompt level of the generative model, encouraging semantic diversity and improving quantization generalisation.  
Instead of using simple text prompt template to generate images, our work learns multiple prompt embeddings per class under diversity and orthogonality constraints, producing semantically correct and visually diverse calibration data.


\section{Method}
\label{sec:method}

\subsection{Overview}

\begin{figure}[t]
    \centering
    \includegraphics[width=0.8\linewidth]{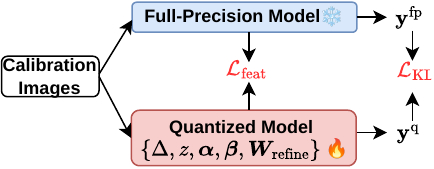}
    \caption{Overview of the end-to-end quantization pipeline}
    \label{fig:pipeline}
\end{figure}
\vspace{-4pt}

As illustrated in Figure~\ref{fig:pipeline}, our framework adopts an end-to-end optimization pipeline for Vision Transformer quantization. Given either real or synthetically generated calibration data, we first initialize quantization parameters including step size, clipping range, and channel-wise rescaling parameters using a small batch of 32 samples. The full network, including both quantization parameters and refinement weights, is then jointly optimized via a distillation loss between the full-precision and quantized models to preserve global correlations and minimize accuracy degradation.

\subsection{Uniform Quantization}
We adopt the standard uniform quantizer, a widely used and hardware-friendly formulation that maps continuous real-valued tensors into discrete integer domains.  
Given an $N$-bit quantizer, a real-valued tensor 
$\mathbf{x} \in \mathbb{R}^d$ (representing either weights or activations) 
is quantized into an integer tensor 
$\mathbf{q} \in \{0, 1, \dots, 2^{N}-1\}^d$ 
using a learnable step size $\Delta$ and zero-point $z \in \mathbb{Z}$ as:
\begin{equation}
    \mathbf{q} = \mathrm{quant}(\mathbf{x})= \mathrm{clip}\!\left(
        \mathrm{round}\!\left(\frac{\mathbf{x}}{\Delta}\right) + z, \,
        0, \, 2^{N}-1
    \right),
    \label{eq:quant}
\end{equation}
where $\mathrm{clip}(\cdot)$ constrains the integer representation to the valid quantization interval.  
The quantized tensor can then be dequantized (reconstructed) back to the approximate full-precision domain as:
\begin{equation}
    \hat{\mathbf{x}} = \mathrm{dequant}(\mathbf{q}) = (\mathbf{q} - z)\Delta.
    \label{eq:dequant}
\end{equation}

The quantization pair $(\Delta, z)$ jointly defines the mapping between the full-precision and integer domains, where $\Delta$ controls the quantization resolution and $z$ determines the offset of the quantization range.  
In effect, a linear layer transformation $\mathbf{Y} = \mathbf{W}\mathbf{X} + \mathbf{b}$, with $\mathbf{W} \in \mathbb{R}^{d_{\text{out}} \times d_{\text{in}}}$ denoting the weight matrix and $\mathbf{X} \in \mathbb{R}^{d_{\text{in}} \times *}$ the input activation, becomes under quantization:
\begin{equation}
    \mathbf{Y} \approx \hat{\mathbf{W}} \cdot \hat{\mathbf{X}} + \mathbf{b},
    \label{eq:quant_layer}
\end{equation}
where $\hat{\mathbf{W}} = \mathrm{dequant}(\mathrm{quant}(\mathbf{W}))$ and 
$\hat{\mathbf{X}} = \mathrm{dequant}(\mathrm{quant}(\mathbf{X}))$ denote the dequantized counterparts of the quantized weights and activations, respectively.  

For activations, a single scalar step size $\Delta_x$ and zero-point $z_x$ are shared across the entire matrix. For the weight matrix $\mathbf{W}$, we employ the commonly used channel-wise quantization \cite{li2023repq}, where each output channel (i.e., each row of $\mathbf{W}$) maintains its own learnable step size $\Delta_{w,i}$ and zero-point $z_{w,i}$.

\noindent\textbf{Initialization.}
The quantization parameters $\Delta$ and $z$ are initialized using a small calibration batch (32 images in our setup) to estimate the activation and weight statistics of each layer.  
Let $x_{\min}$ and $x_{\max}$ denote the lower and upper quantization bounds.  
Instead of directly using the raw minimum and maximum values, which are often affected by rare but large-magnitude outliers in transformer activations, using the robust percentile based range, the step size and zero-point are then initialized as:
\begin{equation}
    \Delta = \frac{ P_{99.9}(\mathbf{x}) -  P_{0.1}(\mathbf{x})}{2^{N} - 1}, 
    \quad
    z = \mathrm{round}\!\left(-\frac{ P_{0.1}(\mathbf{x})}{\Delta}\right),
    \label{eq:init}
\end{equation}
where $P_{0.01}(x)$ and $P_{99.9}(x)$ represents the $0.01$ and $99.9$ percentile value of x, with $0.01$ and $99.9$ empirically chosen for robustness

\noindent\textbf{Differentiable Quantization.}
After initialization, both $\Delta$ and $z$ are treated as learnable parameters during fine-tuning.  
Since the rounding operation in Eq.~\ref{eq:quant} is non-differentiable, we employ the \textit{straight-through estimator} (STE)~\cite{bengio2013estimating} to enable gradient-based optimization:
\begin{equation}
    \mathrm{round_{STE}}(x) = x + (\mathrm{round}(x) - x).\mathrm{detach()},
    \label{eq:ste}
\end{equation}
which copies gradients from the input $x$ to the rounded output during backpropagation, allowing the quantization process to be seamlessly integrated into the end-to-end optimization framework.

\subsection{Channel-Wise Rescaling}

While uniform quantization is conceptually simple, transformer architectures exhibit strong channel-wise variance in activation magnitudes and contain crucial outliers, particularly within attention outputs \cite{li2023repq, xiao2023smoothquant}. 
Therefore, a single global step size $\Delta$ and zero-point $z$ shared across all channels cannot effectively represent both small and large activation channels, leading to severe information loss. 

To mitigate this issue, we use \textbf{channel-wise rescaling and shifting} mechanism inspired by SmoothQuant~\cite{xiao2023smoothquant} and RepQ-ViT~\cite{li2023repq} for each of the input channel, which is applied to each post-LayerNorm fully connected layer.  
Let the input activation be $\mathbf{X} \in \mathbb{R}^{d_{\text{in}} \times B}$, where $d_{\text{in}}$ is the input feature dimension and $B$ is the number of tokens.  
We define two learnable vectors: a scale vector $\boldsymbol{\alpha} \in \mathbb{R}^{d_{\text{in}}}$ and a shift vector $\boldsymbol{\beta} \in \mathbb{R}^{d_{\text{in}}}$, used to normalize activations before quantization:
\begin{equation}
    \mathbf{X}'_{c,:} = \frac{\mathbf{X}_{c,:} - \boldsymbol{\beta}_c}{\boldsymbol{\alpha}_c},
    \label{eq:rescale}
\end{equation}
where the division and subtraction are applied channel-wise.  
Each element $\alpha_c$ in $\boldsymbol{\alpha}$ suppresses channels with large magnitudes and amplifies smaller ones, thereby smoothing the dynamic range of activations across channels.  
Here, $\mathbf{X}_{c,:}$ denotes the $c$-th row of the activation matrix corresponding to the $c$-th input feature channel.

To maintain functional equivalence of the original layer transformation $\mathbf{Y} = \mathbf{W}\mathbf{X} + \mathbf{b}$, the corresponding weight matrix column $\mathbf{W}_{:,c}$ is scaled inversely:
\begin{equation}
    \mathbf{W}_{:,c}' = \mathbf{\alpha}_c \, \mathbf{W}_{:,c},
    \label{eq:rescale_weight}
\end{equation}
where $\mathbf{W}_{:,c}$ denotes the $c$-th column of the weight matrix.  

Finally, the bias term $\mathbf{b} \in \mathbb{R}^{d_{\text{out}}}$ is adjusted accordingly to remain consistent after rescaling and shifting:
\begin{equation}
    \mathbf{b}' = \mathbf{b} + \mathbf{W} \boldsymbol{\beta}.
    \label{eq:rescale_bias}
\end{equation}
This formulation ensures that the affine transformation $\mathbf{Y} = \mathbf{W}\mathbf{X} + \mathbf{b}$ remains functionally equivalent after introducing the per-channel normalization defined in Eq.~\ref{eq:rescale}.

This reparameterization smooths the inter-channel activation landscape by equalizing channel scales, reducing the burden on a single global quantization range and preventing small-magnitude channels from collapsing to zero.
It effectively transforms $(\mathbf{W}, \mathbf{X})$ into $(\mathbf{W}', \mathbf{X}')$ with improved statistical balance and quantization compatibility.  
Making the rescaling and shifting parameters learnable further allows adaptation beyond the calibration batch statistics for more flexible optimization.
This compensation preserves the overall output of the layer while redistributing the quantization difficulty from activations to weights, where quantization tends to be more stable.

\paragraph{Initialization.}
The rescaling and shift parameters are initialized using the same calibration batch (32 images) employed for quantization parameter estimation.  
Following the outlier-robust heuristic of SmoothQuant~\cite{xiao2023smoothquant}, we compute the initialization statistics from the calibration data using percentile-based ranges.  
The shift and scaling factors are then initialized as:
\begin{align}
    \boldsymbol{\beta}_c &= \mathrm{median}(\mathbf{X}_{c,:}), \\
    \boldsymbol{\alpha}_c &= 
    \bigg(\frac{P_{99.9}(\mathbf{X}_{c,:})-P_{0.01}(\mathbf{X}_{c,:}) }{P_{99.9}(\mathbf{W}_{:,c})-P_{0.01}(\mathbf{W}_{:,c})}\bigg)^{0.5}
    \label{eq:init_scale}
\end{align}
This initialization balances the quantization difficulty for activations and weights, providing stable starting values before end-to-end optimization.

Once initialized, both $\boldsymbol{\alpha}$ and $\boldsymbol{\beta}$ become learnable parameters that are jointly optimized together with other quantization parameters (step sizes, zero-points, etc.) during end-to-end fine-tuning. 
In practice, we first initialize $\boldsymbol{\alpha}$ and $\boldsymbol{\beta}$, and then compute the initial step size $\Delta$ and zero-point $z$ of the uniform quantizer based on the shifted and scaled weight/activation statistics. 
All parameters are subsequently updated jointly during training.




\subsection{End-to-End Optimization}

After initialization, all quantization-related parameters are optimized jointly in a differentiable framework, where $\{ \Delta, z, \boldsymbol{\alpha}, \boldsymbol{\beta} \}$ are learnable.
We additionally introduce a weight refinement term $\mathbf{W}_{\text{refine}}$ to allow each quantized weight tensor be slightly refined w.r.t. its frozen full-precision counterpart. 
The refinement term is initialized with zeros to ensure that the quantized weights initially match the baseline model:

\begin{equation}
    \hat{\mathbf{W}_q} = \mathrm{quant}(\mathbf{W}_{\text{fp}} + \mathbf{W}_{\text{refine}})
    \label{eq:w_refine}
\end{equation}

\subsubsection{Loss Function}

No labeled data are used during our optimization process.  
For the same input batch, the Vision Transformer (ViT) is evaluated twice: once using the frozen full-precision model to obtain reference outputs, and once using the quantized model to obtain trainable outputs.  
Let $\{\mathbf{y}_i^{\text{fp}}\}_{i=1}^{B}$ and $\mathbf{y}^{\text{fp}}$ denote the intermediate and final outputs of the full-precision model, respectively,  
$\{\mathbf{y}_i^{\text{q}}\}_{i=1}^{B}$ and $\mathbf{y}^{\text{q}}$ denote the corresponding outputs of the quantized model and $B$ denotes the number of blocks.

The overall loss objective combines the intermediate feature reconstruction loss, the final-layer Kullback–Leibler (KL) divergence loss, and an $\ell_1$ regularization term on the weight refinement parameters $\mathbf{W}_{\text{refine}}$:
\begin{equation}
    \mathcal{L} = 
    \lambda_{\text{feat}} \, \mathcal{L}_{\text{feat}} 
    + \lambda_{\text{KL}} \, \mathcal{L}_{\text{KL}} 
    + \lambda_{\text{reg}} \, \mathcal{L}_{\text{reg}}.
    \label{eq:loss_total}
\end{equation}

\paragraph{Intermediate Feature Reconstruction.}
Intermediate features between corresponding transformer blocks are aligned using a Mean Squared Error (MSE) reconstruction loss:
\begin{equation}
    \mathcal{L}_{\text{feat}} = \sum_{i=1}^{B} \frac{1}{N_i}
    \| \mathbf{y}_i^{\text{fp}} - \mathbf{y}_i^{\text{q}} \|_2^2,
    \label{eq:loss_feat}
\end{equation}
where $N_i$ denotes the number of elements in feature map $\mathbf{y}_i$. 

\paragraph{Final Logit Distillation.}
To match the semantic prediction distributions, we use a temperature scaled KL divergence loss between the full-precision and quantized model logits:
\begin{equation}
    \mathcal{L}_{\text{KL}} = 
    T^2 \cdot 
    \mathrm{KL}\!\left(
        \mathrm{softmax}\!\left(\frac{\mathbf{y}^{\text{fp}}}{\tau}\right)
        \bigg\|
        \mathrm{softmax}\!\left(\frac{\mathbf{y}^{\text{q}}}{\tau}\right)
    \right),
    \label{eq:loss_kl}
\end{equation}
where $\tau=3$ is the temperature factor that smooths the teacher’s output distribution \cite{hinton2015distilling}.

All quantization-related parameters, including step sizes, zero points, rescaling and shifting factors, and refined weights, are jointly optimized via backpropagation using the Adam optimizer. 
We use a learning rate of 1e\text{-}3 for all parameters except the refinement term $\mathbf{W}_{\text{refine}}$, which uses 1e\text{-}4. 
A linear warm-up is applied for the first 5{,}000 iterations, followed by a cosine decay schedule. 
The optimization runs for 24{,}000 iterations with batch size of 32, and converges within about one hour on a single GPU for the ViT-small model.


\subsection{Class-Specific Multi-Prompt Optimization}
\label{sec:multi_prompt_learning}

\begin{figure}[t]
    \centering
    \includegraphics[width=0.99\linewidth]{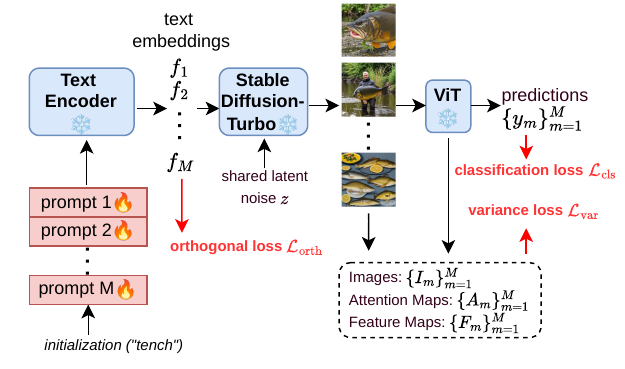}
    \caption{Overview of the data-free multi-prompt learning pipeline. For each class (e.g., “tench”), multiple learned prompts are encoded by a frozen CLIP text encoder and used by Stable Diffusion-Turbo to synthesize diverse images under shared latent noise. A frozen ViT classifier provides supervision via classification loss, while orthogonality and variance losses encourage semantic and visual diversity.}
    \label{fig:datafreepipeline}
\end{figure}

For data-free quantization, we employ Stable Diffusion Turbo~\cite{sauer2024fast} to synthesize calibration images. Instead of relying on a single handcrafted prompt such as “a \texttt{<adjective>} photo of \texttt{<class>}”, we learn multiple prompt embeddings for each of the 1,000 ImageNet classes. 
Specifically, for each class we learn $M$ distinct prompts that are optimized to generate images that are correctly classified as the target class by a pretrained ViT classifier, and also produce diverse image appearances that covers different object layouts, textures and scenes, when conditioned on the same initial noise. 
An overview of the process is illustrated in Fig.~\ref{fig:datafreepipeline}.

\textbf{Prompt initialization.}
Each prompt consists of 20 tokens, where each token embedding has a 1024-dimensional representation, matching the CLIP text encoder used in SD-Turbo.
For a given class, the first 10 tokens are initialized using its textual description (e.g., “tench, tinca tinca”), truncated if longer than 10 tokens. 
The remaining 10 context tokens are initialized randomly with $\ell_2$ norms uniformly sampled in the range $[0.3, 0.4]$, roughly matching the token embedding scale of CLIP.
This initialization provides both semantic grounding and optimization flexibility.

\textbf{Optimization pipeline.} The $20\times1024$ token embeddings are passed through the CLIP text encoder to obtain the text embedding, which is used by SD-Turbo to synthesize images. 
These generated images are then fed into a frozen, pretrained ImageNet ViT classifier to compute the classification loss:
\[
\mathcal{L}_{\text{cls}} = - \log p(y_{\text{cls}} | I_{\text{gen}}),
\]
which encourages each prompt to generate images recognized as its corresponding class.


\textbf{Diversity regularization.}
To promote intra-class diversity among the $M$ learned prompts, we apply complementary regularizers:
\begin{equation}
\begin{split}
\mathcal{L}_{\text{orth}} &= \sum_{i\neq j} | f_i^\top f_j |, \\
\mathcal{L}_{\text{var}} &= -\big[\mathrm{Var}(I_m) + \mathrm{Var}(F_m) + \mathrm{Var}(A_m)\big],
\end{split}
\end{equation}
where $f_i$ are EOS text embeddings, $I_m$ are generated RGB images, 
$F_m$ their ViT features, and $A_m$ the corresponding attention maps.  
The orthogonality term encourages distinct prompt directions, 
while the combined variance term increases diversity in appearance, feature, and spatial attention space.

\textbf{Training stability.}
Since the gradient flow through the prompt–image mapping can be unstable, certain prompts may drift semantically far away from the target class during optimization.
When the classification loss of a prompt exceeds a predefined threshold, we reinitialize its embedding as a randomly weighted average of the remaining well-performing prompts, perturbed with small random noise.
This heuristic stabilizes training and preserves coverage across multiple valid visual modes.

\textbf{Training setup.}
For each class, we optimize $M$ prompt embeddings for 120 iterations using the Adam optimizer with a learning rate of 1e\text{-}3. 
All prompts are jointly updated while sharing the same random latent noise at each iteration, ensuring that diversity originates solely from the learned prompts rather than stochastic initial noise. 
The overall objective is a weighted sum of the classification and diversity related terms:
\begin{align}
\mathcal{L}_{\text{total}} 
&= \lambda_{\text{cls}}\mathcal{L}_{\text{cls}} 
+ \lambda_{\text{orth}}\mathcal{L}_{\text{orth}} 
+ \lambda_{\text{rgb}}\mathcal{L}_{\text{var-rgb}} \nonumber\\
&\quad + \lambda_{\text{feat}}\mathcal{L}_{\text{var-feat}} 
+ \lambda_{\text{attn}}\mathcal{L}_{\text{var-attn}},
\end{align}

where $\lambda_{\text{cls}}{=}1$ and all diversity-related weights ($\lambda_{\text{orth}}$, $\lambda_{\text{rgb}}$, $\lambda_{\text{feat}}$, $\lambda_{\text{attn}}$) are set to $0.1$.






\section{Experiments}
\label{sec:experiments}

\newcommand{\best}[1]{\textbf{#1}}
\newcommand{\second}[1]{\underline{#1}}

\newcommand{\wthree}{W3A3}
\newcommand{\wfour}{W4A4}
\newcommand{\wsix}{W6A6}

We evaluate our method on two fronts. First, we assess end-to-end PTQ on real ImageNet calibration data across multiple model types under various bit settings, and study how performance scales with calibration size. Next, we examine the data-free variant, where calibration samples are generated by Stable Diffusion Turbo guided by learned multi-mode prompts, comparing it against baseline text prompts through both quantitative PTQ accuracy and qualitative analyses with visual and t-SNE comparisons.
All experiments are conducted on a single NVIDIA RTX 6000 Ada Generation GPU. The PTQ process takes approximately 1 hour for ViT-Small and up to 2.5 hours for Swin-Base, while learning the multi-mode prompts requires about 3 minutes per class.

\subsection{Effect of Calibration Set Size}
\label{subsec:scaling}
\begin{figure}[tbh]
    \centering
    \includegraphics[width=0.9\linewidth]{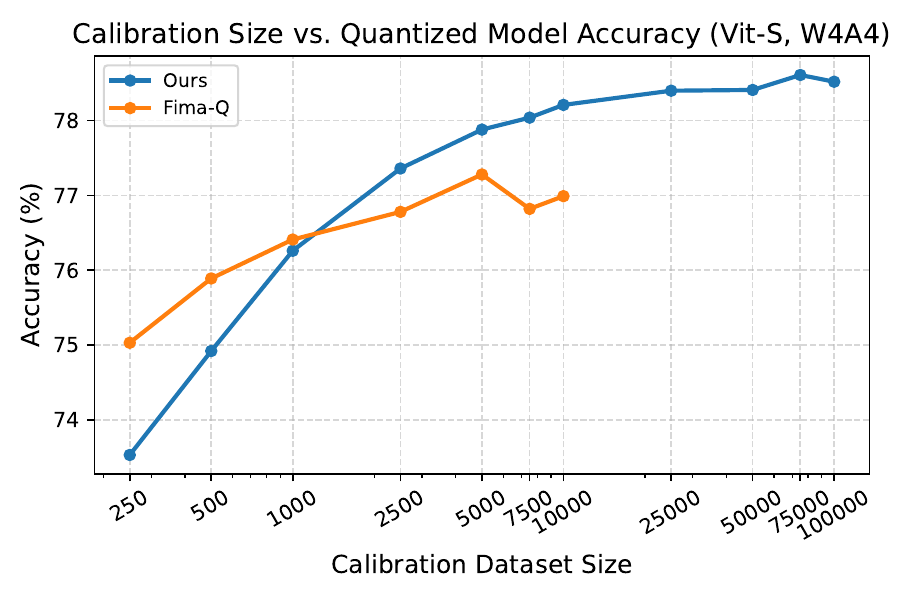}
    \caption{Accuracy vs.~calibration size for ViT-S (W4A4): comparison with FIMA-Q.}
    \label{fig:datascaling_fimaq}
\end{figure}

We study how quantization accuracy scales with the amount of calibration data and compare our end-to-end PTQ with the block-reconstruction method FIMA-Q \cite{wu2025fima}.
As shown in Fig.~\ref{fig:datascaling_fimaq}, using ViT-S under W4A4, our method consistently benefits from larger calibration sets, up to around 10,000 samples, after which the performance gain saturates. FIMA-Q also improves with more data but to a lesser extent, and cannot be evaluated beyond 10,000 samples due to GPU memory limits required for storing intermediate feature maps during reconstruction.

\begin{figure}[tbh]
    \centering
    \includegraphics[width=0.9\linewidth]{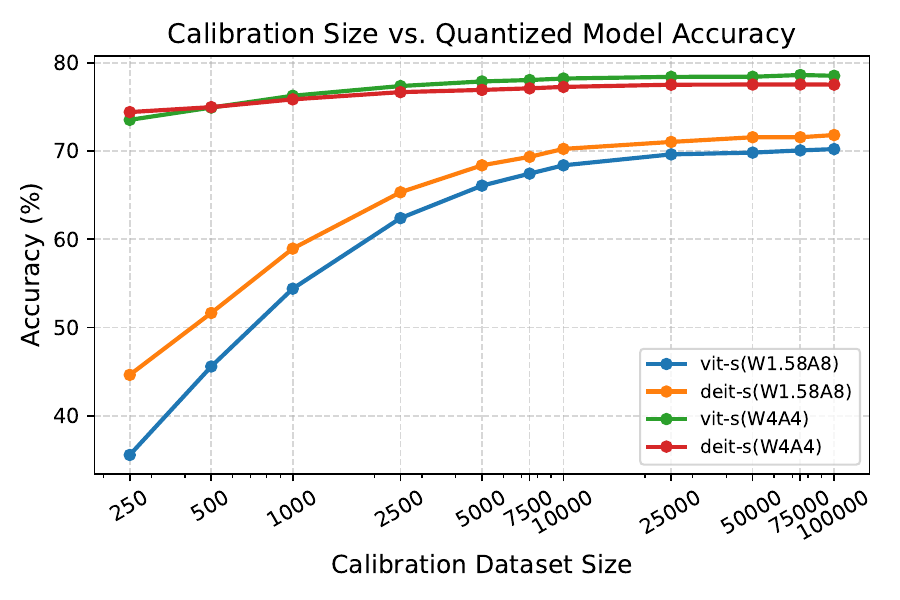}
    \caption{Scaling trends across models and bit settings. Performance gains starts to diminish beyond 10k calibration samples.}
    \label{fig:datascaling_curve}
\end{figure}

We further extend this analysis across ViT-S and DeiT-S under W4A4 and W1.58A8 (Fig.~\ref{fig:datascaling_curve}). Accuracy increases steadily with larger calibration sets, particularly for ultra-low-bit (W1.58A8) settings, but saturates beyond 10,000 samples. Hence, we use 10,000 real images for all subsequent experiments.
Unlike block-wise reconstruction, our flexible end-to-end joint optimization allows cross-block compensation, enabling more effective adaptation across the entire network.

\subsection{Comparison with State-of-the-Art}
\label{subsec:ptq_real}

We evaluate our method against leading PTQ approaches including RepQ-ViT~\cite{li2023repq}, FIMA-Q~\cite{wu2025fima}, and APHQ-ViT~\cite{wu2025aphq} across multiple backbones (ViT, DeiT, Swin) and bit settings (W1.58A8, W3A3, W4A4, W6A6). 
Results are summarized in Tab.~\ref{tab:sota_comparison},
(Real) denotes 10,000 real ImageNet calibration samples, while (Synth) refers to our data-free variant using 100,000 synthetic images generated with learned multi-mode prompts. 

\begin{table}[t]
\centering
\setlength{\tabcolsep}{4pt}
\small
\caption{
Top-1 accuracy (\%) on ImageNet-1K for various backbones and bit-widths.  
Real: 10k real calibration images.  
Synth: 100k synthetic images generated via our learned multi-mode prompts.  
Best results are in \best{bold}, second best in \second{underlined}.
}
\label{tab:sota_comparison}
\begin{tabular}{lcccccc}
\toprule
\multirow{2}{*}{Method} & \multicolumn{2}{c}{ViT} & \multicolumn{2}{c}{DeiT} & \multicolumn{2}{c}{Swin} \\
\cmidrule(lr){2-3}\cmidrule(lr){4-5}\cmidrule(lr){6-7}
 & S & B & S & B & S & B \\
\midrule
Full-Precision &81.39 & 84.54 & 79.85 & 81.80 &83.23 & 85.27 \\
\midrule
\textbf{W1.58A8} \\
ReQ-Vit \cite{li2023repq} &0.12 &0.13 &0.11 & 0.11 & 0.09 & 0.09  \\
FIMA-Q~\cite{wu2025fima}  & 4.84 & 45.55 & 33.93 & 57.75 &  38.13 &  48.10  \\
\textbf{Ours} (Synth) & \second{63.71} & \second{76.58} & \second{67.06} & \second{77.18} & \second{73.06} & \second{75.51} \\
\textbf{Ours} (Real)      & \best{68.45} & \best{78.51} & \best{70.13} & \best{78.07} & \best{76.23} & \best{78.89} \\
\midrule
\textbf{W3A3} \\
ReQ-Vit \cite{li2023repq} &0.47 &0.23 &4.74 &7.70 &1.71 & 1.63 \\ 
FIMA-Q~\cite{wu2025fima} & 64.09 & 77.63 & 69.13 & 76.54 & \second{77.26} & 78.82 \\
APHQ-ViT~\cite{wu2025aphq} & 63.17 & 76.31 & 68.76 & 76.31 & 76.10 & 78.14 \\
\textbf{Ours} (Synth)     & \second{68.46} & \second{78.53} & \second{69.59} & \second{77.54} & 76.90 & \second{78.84} \\
\textbf{Ours} (Real)      & \best{71.89} & \best{79.63} & \best{71.55} & \best{78.02} & \best{78.41} & \best{80.24} \\
\midrule
\textbf{W4A4} \\
RepQ-ViT~\cite{li2023repq} & 65.05 & 68.48 & 69.03 & 75.61 & 79.45 & 78.32 \\
FIMA-Q~\cite{wu2025fima} & 76.68 & 83.04 & 76.87 & 80.33 & \best{81.82} & \best{83.60} \\
APHQ-ViT~\cite{wu2025aphq} & 76.07 & 82.41 & 76.40 & 80.21 & \second{81.81} & 83.42 \\
\textbf{Ours} (Synth)     & \second{77.61} & \second{83.35} & \second{77.16} & \second{80.52} & 81.59 & 83.33 \\
\textbf{Ours} (Real)      & \best{78.35} & \best{83.47} & \best{77.25} & \best{80.78} & 81.68 & \second{83.44} \\
\midrule
\textbf{W6A6} \\
RepQ-ViT~\cite{li2023repq} & 80.43 & 83.62 & 78.90 & 81.27 & \second{82.79} & 84.57 \\
FIMA-Q~\cite{wu2025fima} & 80.64 & \second{84.82} & \best{79.52} & \second{81.74} & \best{83.19} &\best{85.01} \\
\textbf{Ours} (Synth)     & \best{80.98} & \best{84.84} & \second{79.46} & \best{81.74} & 82.91 & 84.70 \\
\textbf{Ours} (Real)      & \second{80.84} & 84.80 & 79.43 & 81.69 & 82.84 & \second{84.72} \\
\bottomrule
\end{tabular}
\vspace{-4pt}
\end{table}

Across ViT, DeiT, and Swin backbones, our method yields substantial improvements over prior post-training quantization approaches, especially in low-bit settings where quantization errors are most severe.
At the extreme W1.58A8 configuration, FIMA-Q either fails completely or suffers severe performance degradation, while our end-to-end PTQ maintains stable and high accuracy, showing strong robustness under aggressive quantization.
The advantage remains clear at W3A3 and W4A4, where globally joint optimization enables cross-block error compensation beyond block-wise reconstruction.
Using synthetic calibration data introduces only minor accuracy drops (typically within 1–2\%), confirming the diversity and representativeness of our learned multi-mode prompts.
For Swin models, the gain is smaller at moderate or high bit-widths, likely due to their local attention and normalization structures that already stabilize quantization.
At W6A6, performance converges across methods as quantization becomes less challenging, yet our approach remains competitive and close to full precision.

\subsection{Data-Free Results and Analyses}
\label{subsec:datafree}
We compare baseline raw text prompts (``a \texttt{<template>} photo of \texttt{<cls>}``) against our learned \textit{multi-mode prompts}. The latter are optimized using classification supervision and diversity regularization, without access to any real calibration data.  

\paragraph{Qualitative diversity and ambiguity resolution.}
Figure~\ref{fig:qualitative_raw_vs_learned} compares synthetic images from both prompt types for representative ImageNet classes.  
For ambiguous categories \textit{kite}, raw prompts tend to collapse into a single dominant interpretation, which may not match the ImageNet definition. In contrast, our learned multi-mode prompts produce semantically correct and visually diverse samples, capturing multiple object configurations, backgrounds, and styles.  
This diversity arises naturally from our prompt-orthogonality and feature-variance regularization, encouraging distinct but class consistent generation modes.





\begin{figure}[tbh]
    \centering
    \includegraphics[width=0.32\linewidth]{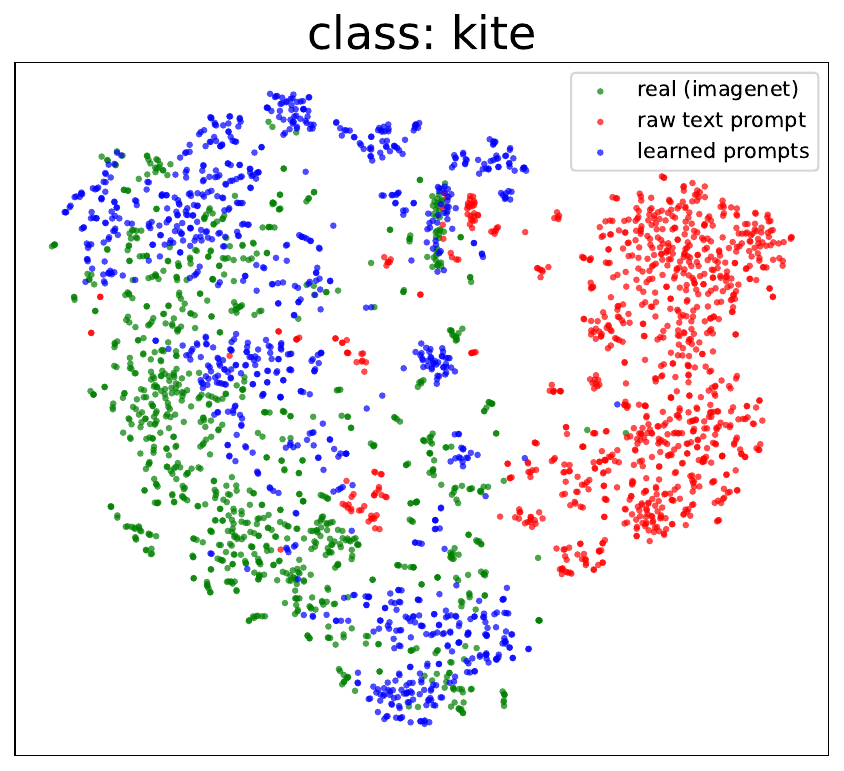}
    \includegraphics[width=0.32\linewidth]{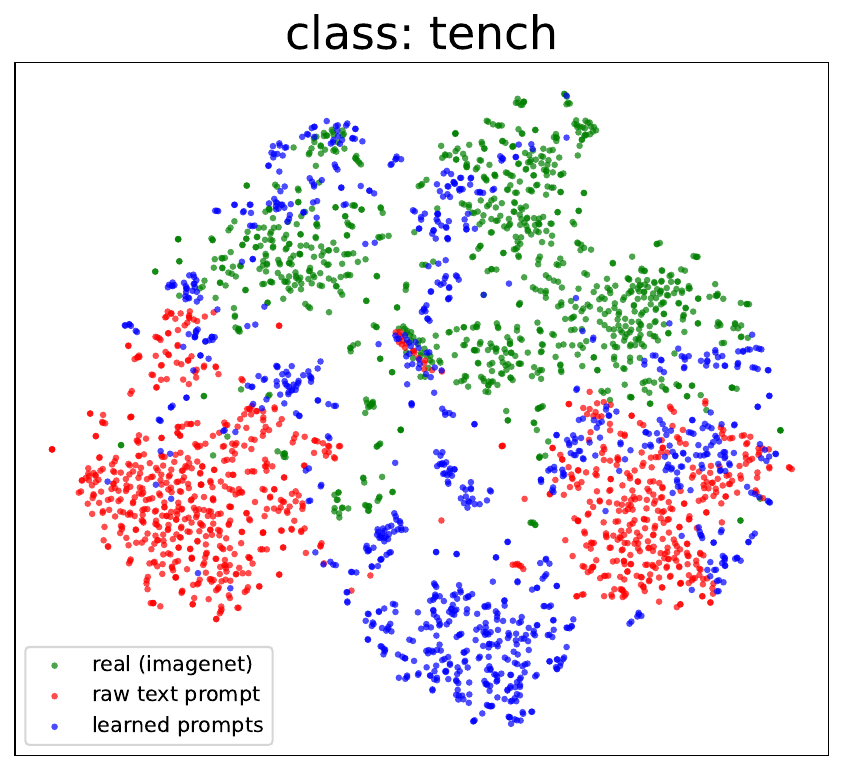}
    \includegraphics[width=0.32\linewidth]{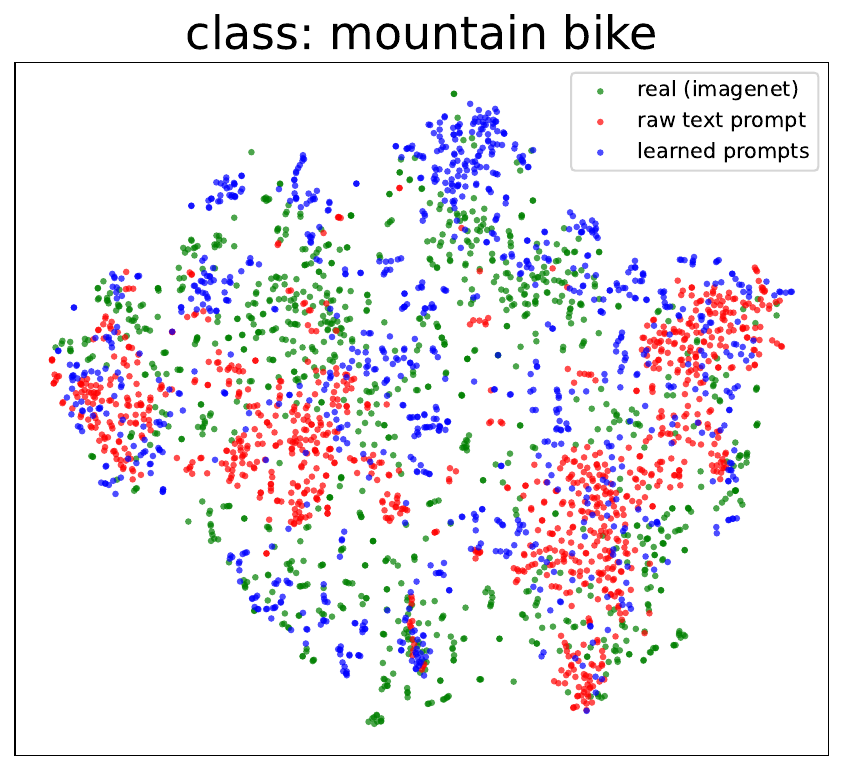}
    \vspace{-5pt}
    \caption{
        \textbf{t-SNE visualization of ViT-Small feature embeddings.}
        Each plot shows 100 samples per class for \textit{kite}, \textit{tench}, and \textit{mountain bike}.  
        \textcolor{green}{\textbf{Green:}} real ImageNet images.  
        \textcolor{red}{\textbf{Red:}} synthetic images from raw text prompts.  
        \textcolor{blue}{\textbf{Blue:}} synthetic images from our learned multi-mode prompts.  
        Our method produces broader, multi-cluster feature distributions closer to the real-image distribution, suggesting improved semantic coverage and feature diversity.
    }
    \label{fig:tsne_raw_vs_learned}
\end{figure}

\paragraph{Feature-level distribution: t-SNE visualization.}
To further analyze the effect of learned prompts, we visualize ViT-Small feature embeddings from synthetic images using t-SNE (Fig.~\ref{fig:tsne_raw_vs_learned}).  
For each class, we extract image token features from the last transformer block for 100 randomly generated samples. Green points correspond to real ImageNet images, red to baseline raw text prompt generations, and blue to our learned multi-mode prompts.  
The plots show that features from our learned prompts are more widely distributed and cluster closer to the real-data manifold, whereas raw text prompt samples form tighter, biased clusters reflecting limited semantic coverage.  
This confirms that multi-mode prompt learning enhances feature diversity and domain alignment, leading to better calibration quality in downstream quantization.

\begin{figure}[tbh]
    \centering
    \includegraphics[width=0.9\linewidth]{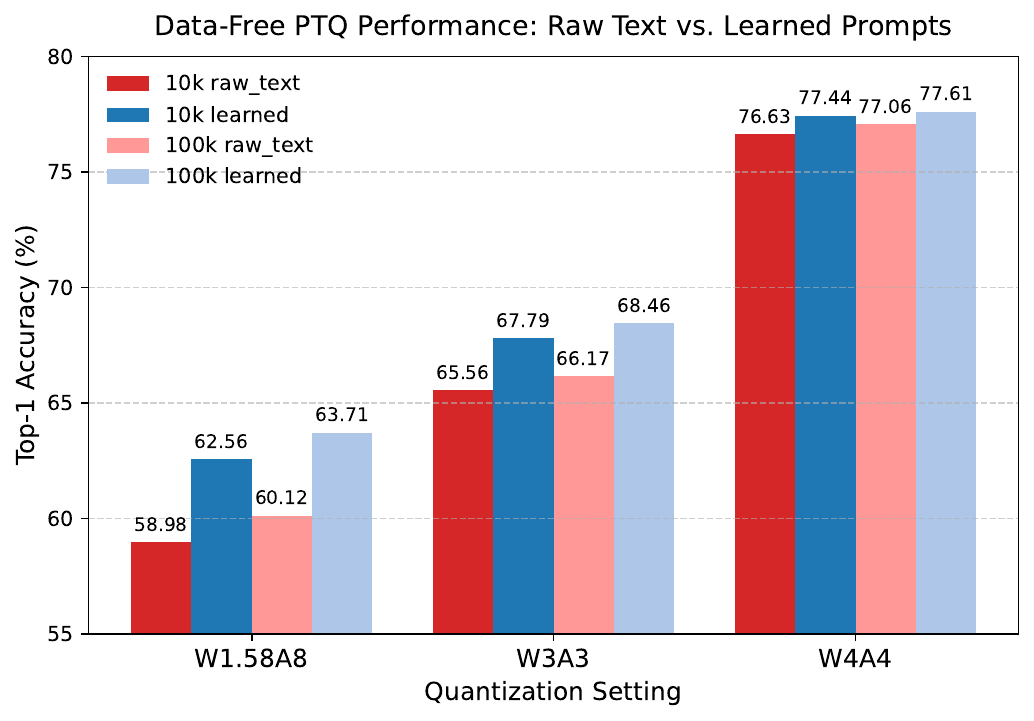}
    \vspace{-3pt}
    \caption{
        \textbf{PTQ accuracy comparison between raw and learned prompts.}
        Each group shows top-1 accuracy (\%) on ImageNet for ViT-Small under three bit-width settings (W1.58A8, W3A3, W4A4). 
        Learned prompts consistently outperform raw text prompts across all calibration sizes. 
        The gain is most pronounced in low-bit regimes (e.g., +3.6\% at W1.58A8), 
        indicating that improved semantic coverage and feature diversity from multi-mode prompt learning 
        lead to higher-quality synthetic calibration data.
    }
    \label{fig:compare_raw_learned_ptq}
\end{figure}

\paragraph{Prompt quality impacts PTQ performance.}
To quantify the benefit of our learned prompts for data-free PTQ, 
we evaluate the top-1 accuracy of ViT-Small under three representative bit-width settings: 
W1.58A8, W3A3, and W4A4. 
Each model is calibrated using either \textit{raw text prompts} or our \textit{learned multi-mode prompts}, 
with 10k and 100k synthetic samples, respectively. 
Results are summarized in Fig.~\ref{fig:compare_raw_learned_ptq}.

Learned multi-mode prompts yield consistent accuracy improvements over raw text prompts across all quantization levels.  
The benefit is especially evident under aggressive quantization (W1.58A8 and W3A3), 
where calibration sensitivity is higher and richer synthetic data helps preserve model fidelity.  
At moderate bit-widths (W4A4), the accuracy gap narrows as quantization becomes less error prone, 
but learned prompts still provide small gains.  
These results validate that our data-free prompt learning strategy 
not only improves the realism and diversity of generated samples 
(Fig.~\ref{fig:qualitative_raw_vs_learned}) 
but also translates directly into PTQ performance benefits.
\section{Conclusion}
We proposed an end-to-end post-training quantization framework that jointly optimizes quantization parameters, channel-wise rescaling and weight refinement.
For the data-free setting, we introduced a multi-prompt learning strategy that generates diverse and semantically aligned synthetic samples for each class using only classification signals, without relying on real images.
Our method achieves state-of-the-art accuracy across multiple Vision Transformer backbones under low-bit settings, effectively narrowing the gap between data-free and real-data calibration.
The framework requires only moderate computational cost and consistently performs well across different Vision Transformer architectures.

\clearpage

{
    \small
    \bibliographystyle{ieeenat_fullname}
    \bibliography{main}
}


\end{document}